\documentclass[sn-nature]{sn-jnl}

\usepackage[utf8]{inputenc} 
\usepackage[T1]{fontenc}    
\usepackage{multirow}%
\usepackage{amsmath,amssymb,amsfonts}%
\usepackage{amsthm}%
\usepackage{mathrsfs}%
\usepackage[title]{appendix}%
\usepackage{xcolor}%
\usepackage{textcomp}%
\usepackage{manyfoot}%
\usepackage{booktabs}%
\usepackage{algorithm}%
\usepackage{algorithmicx}%
\usepackage{algpseudocode}%
\usepackage{listings}%
\usepackage[T1]{fontenc}
\usepackage{cleveref}
\usepackage{xcolor}
\usepackage{url}
\usepackage[inkscapeformat=png]{svg}
\usepackage{fontawesome5}
\usepackage{mdframed}
\usepackage{siunitx}
\usepackage{graphicx}%
\usepackage{doi}
\usepackage{natbib}
\usepackage{capt-of}
\usepackage{setspace}
\newif\ifShowReviewMarks
\ShowReviewMarksfalse

\newcommand{\ReviewHighlight}[1]{
\ifShowReviewMarks
    \unskip\textcolor{red}{#1}\ignorespaces
\else
    \unskip\textcolor{black}{#1}\ignorespaces
\fi}

\begin{document}

\title[Deep learning for proximal caries detection]{AI-Dentify: Deep learning for proximal caries detection on bitewing x-ray - HUNT4 Oral Health Study}


\author*[1]{\fnm{Javier} \sur{Pérez de Frutos}}\email{javier.perezdefrutos@sintef.no}
\author[1]{\fnm{Ragnhild} \sur{Holden Helland}}
\author[2]{\fnm{Shreya} \sur{Desai}}
\author[4,5]{\fnm{Line Cathrine} \sur{Nymoen}}
\author[1]{\fnm{Thomas} \sur{Langø}}
\author[3]{\fnm{Theodor} \sur{Remman}}
\author[4,5]{\fnm{Abhijit} \sur{Sen}}
\affil[1]{\orgdiv{Department of Health Research}, \orgname{SINTEF Digital}, \orgaddress{\street{Professor Brochs gate 2}, \city{Trondheim}, \postcode{7030}, \country{Norway}}}
\affil[2]{\orgname{Boneprox A.B.}, \orgaddress{\city{Gothenburg}, \country{Sweden}}}
\affil[3]{\orgname{Boneprox A.S.}, \orgaddress{\city{Tønsberg}, \country{Norway}}}
\affil[4]{\orgdiv{Department of public Health and Nursing}, \orgname{Norwegian University of Science and Technology}, \orgaddress{\city{Trondheim}, \country{Norway}}}
\affil[5]{\orgname{\ReviewHighlight{Kompetansesenteret Tannhelse Midt (TkMidt)}}, \orgaddress{\city{Trondheim}, \country{Norway}}}




\abstract{
\textbf{Background:} Dental caries diagnosis requires the manual inspection of diagnostic bitewing images of the patient, followed by a visual inspection and probing of the identified dental pieces with potential lesions. Yet the use of artificial intelligence, and in particular deep-learning, has the potential to aid in the diagnosis by providing a quick and informative analysis of the bitewing images.

\textbf{Methods:} A dataset of 13,887 bitewings from the HUNT4 Oral Health Study were annotated individually by six different experts, and used to train three different object detection deep-learning architectures: RetinaNet (ResNet50), YOLOv5 (M size), and EfficientDet (D0 and D1 sizes). A consensus dataset of 197 images, annotated jointly by the same six \ReviewHighlight{dental clinicians}, was used for evaluation. A five-fold cross validation scheme was used to evaluate the performance of the AI models.

\textbf{Results:} the trained models show an increase in average precision and F1-score, and decrease of false negative rate, with respect to the dental clinicians. \ReviewHighlight{When compared against the dental clinicians, the }YOLOv5 \ReviewHighlight{model }shows the largest improvement, reporting 0.647 mean average precision, 0.548 mean F1-score, and 0.149 mean false negative rate. Whereas the best annotators on each of these metrics reported 0.299, 0.495, and 0.164 respectively.

\textbf{Conclusion:} Deep-learning models have shown the potential to assist dental professionals in the diagnosis of caries. Yet, the task remains challenging due to the artifacts natural to the \ReviewHighlight{bitewing images}.
}




\maketitle

\section{Introduction}
\label{sec:introduction}

As reported in the WHO Global Oral Health Status Report in 2022 \cite{WHOGlobalOralHealth2022}, globally 3.5 billion people are afflicted by some form of oral disease, and 2 billion suffer from caries in permanent teeth. Furthermore, untreated dental caries in permanent teeth is the most common dental health condition. Diagnosis of such lesions requires both the inspection of clinical images e.g., X-ray (bi-dimensional images) or cone beam computed tomography (tri-dimensional images), as well as the visual examination and probing of the affected tooth or teeth. This procedure is time consuming, and requires a high level experience when analysing the clinical images. The two main image modalities used to assist and support the examination of caries are bitewing (BW) and panoramic radiography (OPG) \cite{Schwendicke2016, Schwendicke2020ConventionalRadiography}. Caries, particularly proximal caries, a type of carious lesion located on the surfaces between adjacent teeth, are difficult to detect manually or visually (i.e. using radiographic X-ray images) due to artifacts. Also,  poor angulation can hinder the correct identification of the lesions or even occlude lesser grade caries. 

Since 2008, the research on the application of artificial intelligence (AI) and, more specifically, deep learning (DL) convolutional neural networks (CNN) models for the analysis of dental has noticeably increased \cite{Devito2008, Berdouses2015, Singh2017, Hwang2019, PradosPrivado2020, Schwendicke2019, Choi2018, Mayank2017, JaeHong2018, Shinae2021}. However, research on this field is still limited compared to other clinical areas. Data availability and reliable annotations \cite{PradosPrivado2020, Shinae2021} are the main bottlenecks in the development of machine learning (ML) methods in dentistry. A large portion of the published work uses a dataset of fewer than $300$ images, only few studies have access to large datasets \cite{PradosPrivado2020} with more than $1,000$ images like \cite{Mayank2017, Cantu2020, Park2022}. Of these publications, the work presented in \cite{Devito2008, Berdouses2015, Singh2017, Mayank2017, Park2022} focus on object detection, which is the scope of the present study. Object detection or object recognition refers to the task of localising and classifying objects in a picture \cite{Godfellow2016}. The localisation is usually marked using axis-aligned bounding boxes, surrounding the outermost boundary of the item of interest.

In Devito~\textit{et~al.,}~\cite{Devito2008}, a multi-layer perceptron with 51 artificial neurons (25 in the input layer, 25 in the hidden layer, and one in the output layer) is used to detect proximal caries on BW images, using a dataset of 160 images annotated by 25 experts. Whereas in Srivastava~\textit{et~al.,}~\cite{Mayank2017}, a caries detector built using a tailor designed fully connected neural network was trained with $3,000$ annotated BW images. In Singh~\textit{et~al.,}~\cite{Singh2017}, hand-crafted features for X-ray images are built using Radon and discrete cosine transformations, and further classified using an ensemble of ML techniques such as random forest. Park~\textit{et~al.,}~\cite{Park2022} proposed an ensemble of U-Net and Fast R-CNN for caries detection in colour image, trained with $2,348$ RGB intraoral photographic images. Even though, the work done by Cantu~\textit{et~al.}~\cite{Cantu2020} focuses on image segmentation, it is worth mentioning because of the dataset used: $3,686$ BW images, with caries segmentation annotations, to train a U-Net model for segmentation.


\subsection{Study goals}
In this study we compare three state-of-the-art deep learning architectures for object detection on the task of proximal caries detection, namely RetinaNet, YOLOv5, and EfficientDet. By using an extensive and annotated dataset, we hypothesised that AI object detection models can perform in equal or better terms than dental clinicians. Hence, in this study we trained the aforementioned architectures in detection and classification of enamel caries, dentine caries, and secondary lesions, in BW images. Then, the models were compared to human annotators in order to test our hypothesis. In addition, a novel processing pipeline for merging multi-observer object detection annotations, based on Gaussian Mixture Models, is proposed. 

\section{Methods}
\label{sec:methods}

\subsection{Dataset}
\label{subsec:dataset}
The bitewing images used in this study were collected as part of the HUNT4 Oral Health Study \ReviewHighlight{on the prevalence of periodontitis in a Norwegian population}, a sub-study of the fourth phase of the HUNT study~\cite{HUNTStudyPaper}. The HUNT4 Oral Health Study is a collaborative study between several Norwegian institutes including: the HUNT research centre, the \ReviewHighlight{Kompetansesenteret Tannhelse Midt (TkMidt)}, the Norwegian University of Technology (NTNU), the University of Oslo (UiO), the Tannhelsetjenestens Kompetansesenter Øst (TkØ), and the Norwegian National Centre for Ageing and Health.

The data collected consisted of clinical and radiographic oral examination, which took place between 2017 and 2019. A total of $7,347$ participants were invited to participate in the study, out of a population of $137,233$ people (2017) \cite{HUNT4Paper}. Only \ReviewHighlight{$4,933$ } participants where included in the Oral Health survey study, \ReviewHighlight{out of which, $4,913$ completed both clinical and radiographic examination \cite{HUNT4Paper,HUNT4DataAnalysis}. }A total of $19,210$ BW and \ReviewHighlight{$4,871$ }OPG images where collected from the participants. \ReviewHighlight{The demographics of the dataset showed a distribution of $2759$ ($56\%$) female and $2174$ ($44\%$) male participants, with ages ranging from $19$ to $94$ years ($51.8\pm16.6$ years on average)~\cite{HUNT4Paper}}. For this study, only the BW images were considered.

\begin{figure}[ht]
    {\centering
    \includegraphics[width=0.9\linewidth]{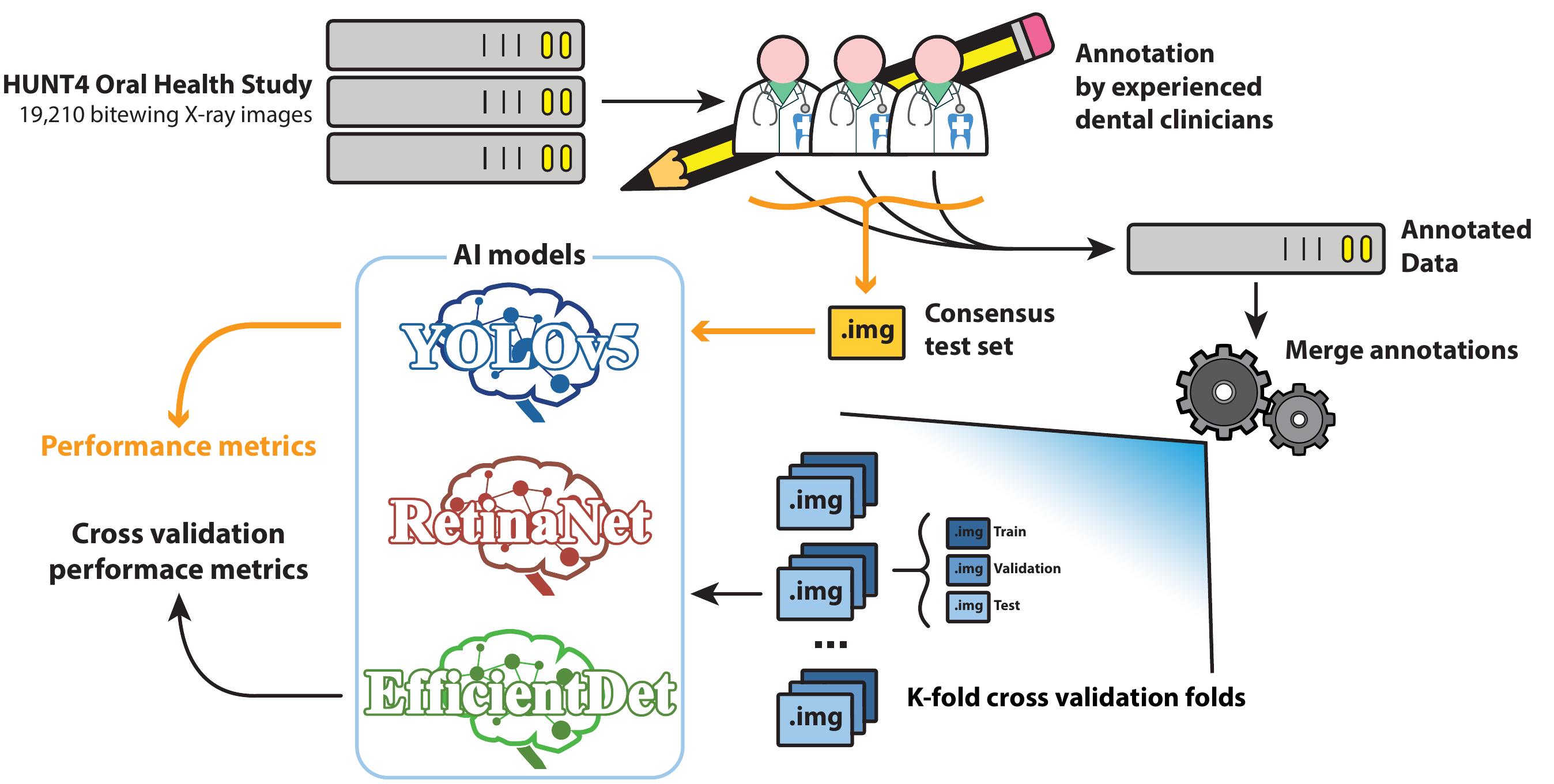}}
    
    \ReviewHighlight{{\setstretch{0.1}\scriptsize The HUNT4 Oral Health Study bitewings were stored on a dedicated server, and made available to the expert dentist and dental hygienists for annotation, resulting in the annotated data and the consensus test set. The resulting annotations were merged to build the  datasets used in this study. The training dataset was further split following a K-fold (in this study $K=5$) cross-validation (CV), and pre-processed. The AI models were trained and evaluated on both the CV test set and the consensus test set.}}
    \caption{\ReviewHighlight{Data workflow}}
    \label{fig:workflow}
\end{figure}

The following subsections will further describe the steps of the workflow followed in the present study, which is depicted in \Cref{fig:workflow}. 

\subsection{Data annotation}
\label{subsec:annotation}
The data was annotated by six dental clinicians with extensive experience in the diagnosis of proximal caries, using the open-source annotation tool AnnotationWeb~\cite{annotationweb}. The caries were classified in five different categories shown in \Cref{tab:labels_definitions}. Further details of the annotation procedure can be found in Section 1 of the Additional Materials 1.

\begin{table}[ht]
    \centering
    \caption{Definition of the classes used to annotate the dataset}
    \begin{tabular}{rl}
        \toprule
        Label name & Description \\
        \midrule
        Grade 1 & Radiolucent in outer half of the enamel \cite{Westberg2010, Hansson2012} \\
        Grade 2 & Radiolucent in the inner half of the enamel, but not in the dentine \cite{Westberg2010, Hansson2012} \\
        Grade 3 & Radiolucent in the outer third of the dentine \cite{Westberg2010, Hansson2012} \\
        Grade 4 & Radiolucent in $2/3$ of the dentine \cite{Westberg2010, Hansson2012} \\
        Grade 5 & Radiolucent in the inner third of the dentine \cite{Westberg2010, Hansson2012} \\
        Secondary lesion & Caries related to \ReviewHighlight{sealants or restorations} \\
        Unknown grade & Caries whose grade cannot be clearly identified \\
        \bottomrule
    \end{tabular}
    \label{tab:labels_definitions}
\end{table}

 To clean the annotations so as to get a ground truth to train the AI models, a novel object detection multi-observers annotations combination strategy was envisioned for this project. First, the annotated bounding boxes were grouped based on the intersection over union (IoU) score, a metric which describes how well the boxes overlap. Then, a Gaussian distribution was fitted to each bounding box in the group, along the vertical and horizontal axes. A mixture density function (MDF) of a Gaussian Mixture Model in which all distributions have the same weight, was obtained by combining the probability density functions of the fitted Gaussian distributions.The common bounding box was then obtained from the MDF given a probability threshold ($p$), as detailed in Algorithm 2, in the Additional Materials 1. Alternatively, the non-maximum suppression (NMS) algorithm can be used to find the best fitting bounding box. However, since all the annotations had the same level of confidence, unlike the predictions done by an AI model, NMS will be biased towards the first bounding box selected as a reference. Lastly, the label of the common bonding box was determined based on the most voted class among the bounding boxes in the group. In case of tie, the most severe class was chosen e.g., dentine caries over enamel caries.

\begin{figure}[h!]
    \centering
    \includegraphics[width=0.8\textwidth]{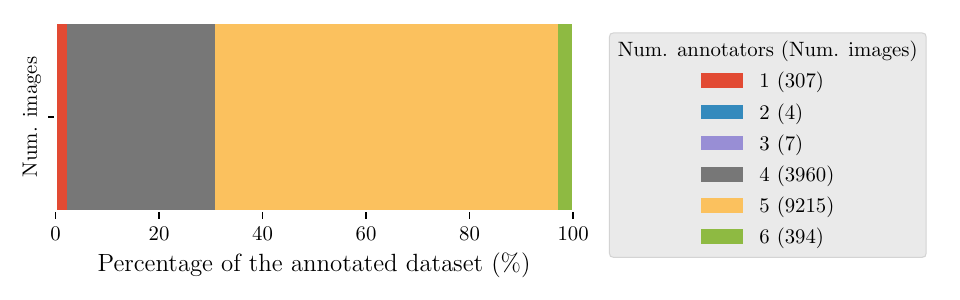}
    \caption{Distribution of annotated images in the annotated dataset. In the legend, the number of annotated images for each interval is shown within brackets.}
    \label{fig:annotated_images}
\end{figure}

A total of $13,887$ images were annotated by one to six of the dental clinicians (see \Cref{fig:annotated_images}), having a total of $13,585$ images annotated by more than one \ReviewHighlight{dental clinician}. 

\begin{figure}[h!]
    \centering
    \includegraphics[width=\linewidth]{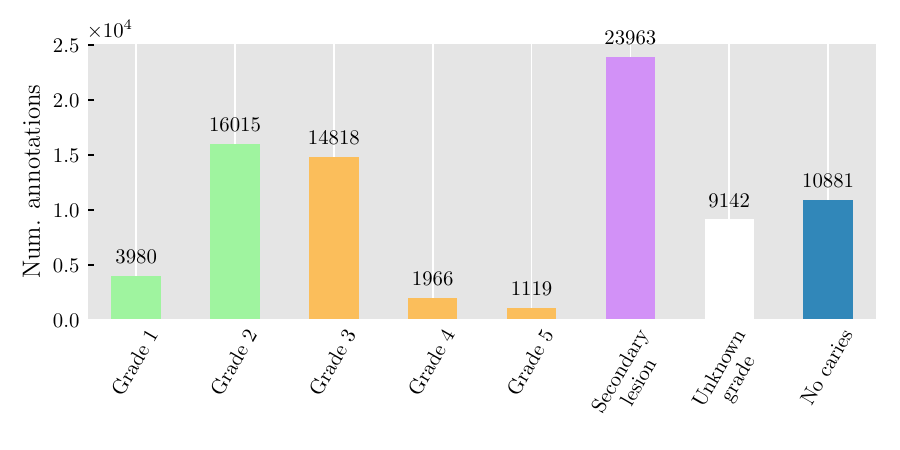}
    \caption{Distribution of annotations in the dataset annotated by the six dental clinicians. Enamel proximal caries (Grades 1 and 2, total $19,995$ annotations) are pictured in light green, dentine lesions (Grade 3 to 5, total $17,903$ annotations) are in orange, secondary lesions are depicted in pink, and caries of uncertain grade have been highlighted in grey. Image free of lesions (No caries) are shown in dark blue, here the number of annotations matches the number of images.}
    \label{fig:distribution_annotations}
\end{figure}

The distribution of labels in Figure \ref{fig:distribution_annotations} shows a higher volume of secondary lesions than all the other grades. After discussion with the dental clinicians, it was agreed to merge the grade one and two under the label of "enamel caries", and grades three to five under the group of "dentine caries". Secondary caries and unknown grade groups were kept as separate label groups.

In addition, $197$ images were annotated by consensus agreement among all the expert annotators, so as to build a test set for evaluation purposes. To create this dataset, hereafter consensus test set, all annotators (dental clinicians) were brought together in the same room \ReviewHighlight{and }agreement \ReviewHighlight{was achieved }by consensus on the annotation of the images. The images in the consensus test set had previously been annotated by all annotators individually, with a considerable time gap between the individual annotations and the creation of the consensus agreement annotations, so that the annotations could be considered independent of each other.


\subsection{Object detection models}
\label{subsec:architectures}
Three state of the art object detection architectures were evaluated for caries detection: RetinaNet (Keras implementation)~\cite{Tsung2017RetinaNet} (ResNet50 backbone), YOLOv5~\cite{Jocher2022YOLOv5} (size M), and EfficientDet~\cite{Mingxing2020EfficientDet} (pretrained D0 and D1). All the models used transfer learning, which is a common strategy when adapting object detection models to a particular dataset, by loading the weights of \ReviewHighlight{models trained }on a larger dataset set e.g., ImageNet or COCO datasets. RetinaNet was initialised with the weights of ResNet50 trained on ImageNet dataset, YOLOv5 loaded the weights pretrained on COCO dataset (provided in the original repository \url{https://github.com/ultralytics/yolov5}), and EfficientDet pretrained weights were obtained from \url{https://github.com/rwightman/efficientdet-pytorch}. For better comparison of the architectures, \Cref{tab:num_params} shows the number of parameters for each architecture. Due to time restrictions, not all the versions of YOLOv5 and EfficientDet are included in the current results. The pre-processing and post-processing were \ReviewHighlight{kept }the same for all models and experiments. Preliminary experiments were conducted with the contrast limited adaptive histogram equalization (CLAHE) method, inspired by Georgieva et al. \cite{Georgieva2017AnEnhancement}, but these experiments were eliminated before the final round of cross-validation because they did not lead to any improvement in the scores. Only  \ReviewHighlight{intensity standardisation to the range $[0, 1]$, and }horizontal and vertical flipping were used to augment the training dataset, both being applied with a probability of $0.5$.

\begin{table}[ht]
    \centering
    \caption{Number of parameters of each architecture.}
    \begin{tabular}{rcl}
        \toprule
        Architecture & Number of parameters (millions) \\
        \midrule
        YOLOv5 M & $21.2$M\\
        \midrule
        RetinaNet (ResNet50) & $36.4$M\\
        \midrule
        EfficientDet D0 & $3.9$M\\
        EfficientDet D1 M & $6.6$M\\
        \bottomrule
    \end{tabular}
    \label{tab:num_params}
\end{table}

The training was done on a dedicated server running Ubuntu 20.04. The machine featured a NVidia Quadro RTX 5000 GPU with 16~GB VRAM, a Intel Core i7-9700 CPU, 32~GB RAM, 1 TB SSD, and 8~TB HDD. \ReviewHighlight{The training parameters for each model are summarised in \cref{tab:train_params}. In the case of YOLOv5 and RetinaNet, the learning rate was monitored using a learning rate scheduler. For YOLOv5, the OneCycleLR scheduler from PyTorch was used with a maximum learning rate of $10^{-1}$. Whereas for RetinaNet, a step learning rate scheduler was used with the patience set to 10 epochs. For both RetinaNet and EfficientDet, an early stopper was used to prevent overfitting. A patience of 20 and minimum loss increment of $5\times10^{-3}$ was configure for RetinaNet, and a patience of 50 and minimum increment of $10^{-1}$ for EfficientDet.}
\begin{table}[ht]
    \centering
    \caption{Training parameters for each model.}
        \begin{tabular}{rcccccll}
            \toprule
            Architecture & Batch size & Learning rate & Max. epochs & Optimiser & Framework\\
            \midrule
            YOLOv5 M & 8 & $10^{-2}$ &  180 &  SGD & PyTorch\\
            \midrule
            RetinaNet & 4 & $10^{-5}$ &  $200$ &  Adam &  Keras\\
            \midrule
            EfficientDet D0 & 8 & $10^{-4}$ & $10^6$ &  AdamW &  PyTorch\\
            EfficientDet D1 & 8 & $10^{-4}$ &  $10^6$ &  AdamW &  PyTorch\\
            \bottomrule
        \end{tabular}  
    \label{tab:train_params}
\end{table}


\subsection{Validation protocol}
\label{subsec:validation-protocol}
After removing the images rejected by the annotators ($980$), images with unknown grade annotations ($4,565$), and those in the consensus test dataset ($197$), the remaining $8,342$ images were split into five folds to perform a cross-validation (CV) study. Random sampling without replacement was used to build the folds. The cross-validation training and evaluation was performed with a three-way-split, i.e. for each iteration, three folds were used for training, one fold was used for validation during training, to avoid overfitting; and the final fold was kept aside as a test set for the final performance evaluation. 

To test our hypothesis on the performance of AI models, both the trained models in each fold and the annotators were evaluated against the consensus test set. However, due to time constraints, one of the annotators did not complete the individual annotation task, missing one image, and thus the resulting metrics of this annotators are not strictly comparable to those of the models and other annotators. This annotator is marked with an asterisk (*) in \cref{tab:results_consensus_annotators_confint}.

\subsection{Performance evaluation}
\label{subsec:performance-evaluation}
As aforementioned, the models described in \Cref{subsec:architectures} and the annotators were evaluated on the consensus test set. The metrics used in the evaluation were the standard metrics for evaluating object detection model performance: average precision (AP) for each of the classes, the mean average precision (mAP) across classes, the F1-score (F1) for each of the classes, the mean F1-score (mF1), as a surrogate for the recall and precision, the false negative rate (FNR) for each class, and the average across classes (mFNR). These three metrics are in the range $[0, 1]$.

Bootstrap confidence intervals ($95\%$) were computed for the test results of both the models and the annotators, to compare the performance of these. \ReviewHighlight{For this comparison, the models trained on the 5\textsuperscript{th}~fold were used}. The intervals were computed using the bias-corrected and accelerated bootstrap algorithm \cite{boostrap1997}, with $1,000$ iterations for confidence interval. Significance in score differences between annotators and models were determined based on overlap of the confidence intervals. 

\section{Results}
\label{sec:results}

The evaluation results on the consensus test set for the five-fold cross-validation can be found in \Cref{tab:results_AP_consensus_summary,tab:results_F1_consensus_summary,tab:results_FNR_consensus_summary} \ReviewHighlight{(the results per fold can be found in Table 1 in Section 4 of the Additional Material 1)}. The metrics were computed using the PASCAL VOC metrics implemented in \cite{padilla2021}, with an IoU thresholds of $0.3$. The threshold was deemed an adequate trade-off between precision and recall for the current application, where detection of potential caries is preferred. The YOLOv5 model reached the highest AP scores for all classes, as well as the highest F1-scores for two out of three classes, and the lowest FNR for all classes.

\begin{table}[!ht]
    \centering
    \caption{Average precision (AP) results (mean and standard deviation) of the five-fold cross-validation evaluated on the consensus test set. The best metrics are highlighted in bold.}
    \label{tab:results_AP_consensus_summary}
        \begin{tabular}{rcccc}
            \toprule
             Model &  Enamel caries & Dentine caries &  Secondary lesion & mAP \\
            \midrule
            YOLOv5 M & \textbf{0.597 $\pm$ 0.034} & \textbf{0.622 $\pm$ 0.029} & \textbf{0.681 $\pm$ 0.022} & \textbf{0.633 $\pm$ 0.025}\\ 
            \midrule
            RetinaNet & 0.371 $\pm$ 0.023 & 0.333 $\pm$ 0.024 & 0.412 $\pm$ 0.034 & 0.372 $\pm$ 0.017\\
            \midrule
            EfficientDet D0 & 0.310 $\pm$ 0.048 & 0.278 $\pm$ 0.017 & 0.267 $\pm$ 0.017 & 0.285 $\pm$ 0.012\\ 
            EfficientDet D1 & 0.340 $\pm$ 0.029 & 0.387 $\pm$ 0.032 & 0.377 $\pm$ 0.028 & 0.368 $\pm$ 0.027\\ 
            \bottomrule
        \end{tabular}
\end{table}

\begin{table}[!ht]
    \centering
    \caption{F1-score results (mean and standard deviation) of the five-fold cross-validation evaluated on the consensus test set. The best metrics are highlighted in bold.}
    \label{tab:results_F1_consensus_summary}
        \begin{tabular}{rcccc}
        \toprule
         Model &  Enamel caries & Dentine caries &  Secondary lesion & mF1 \\
        \midrule
        YOLOv5 M & 0.513 $\pm$ 0.011 &  \textbf{0.588 $\pm$ 0.019} & \textbf{0.563 $\pm$ 0.029}  & \textbf{0.555 $\pm$ 0.011} \\
        \midrule
        RetinaNet & 0.234 $\pm$ 0.032 & 0.312 $\pm$ 0.015 & 0.228 $\pm$ 0.027 & 0.258 $\pm$ 0.013\\ 
        \midrule
        EfficientDet D0 & 0.465 $\pm$ 0.055 & 0.459 $\pm$ 0.021 & 0.444 $\pm$ 0.006 & 0.456 $\pm$ 0.017\\ 
        EfficientDet D1 & \textbf{0.533 $\pm$ 0.021} & 0.561 $\pm$ 0.020 & 0.507 $\pm$ 0.028 & 0.534 $\pm$ 0.019\\ 
        \bottomrule
        \end{tabular}
\end{table}

\begin{table}[!ht]
\centering
\caption{False negative rate (FNR) results (mean and standard deviation) of the five-fold cross-validation evaluated on the consensus test set. The best metrics are highlighted in bold.}
\label{tab:results_FNR_consensus_summary}
    \begin{tabular}{rcccc}
        \toprule
         Model &  Enamel caries & Dentine caries &  Secondary lesion & mFNR \\
        \midrule
        YOLOv5 M & \textbf{0.153 $\pm$ 0.021} & \textbf{0.215 $\pm$ 0.032} & \textbf{0.160 $\pm$ 0.035} & \textbf{0.176 $\pm$ 0.025}\\ 
        \midrule
        RetinaNet & 0.185 $\pm$ 0.060 & 0.364 $\pm$ 0.044 & 0.200 $\pm$ 0.045 & 0.250 $\pm$ 0.031\\  
        \midrule
        EfficientDet D0 & 0.606 $\pm$ 0.060 & 0.636 $\pm$ 0.020 & 0.533 $\pm$ 0.020 & 0.592 $\pm$ 0.025\\ 
        EfficientDet D1 & 0.479 $\pm$ 0.018 & 0.524 $\pm$ 0.020 & 0.459 $\pm$ 0.021 & 0.487 $\pm$ 0.011\\ 
        \bottomrule
    \end{tabular}
\end{table}

\begin{figure}[ht]
    \begin{minipage}{\textwidth}
    \centering
    \includegraphics[width=0.9\textwidth]{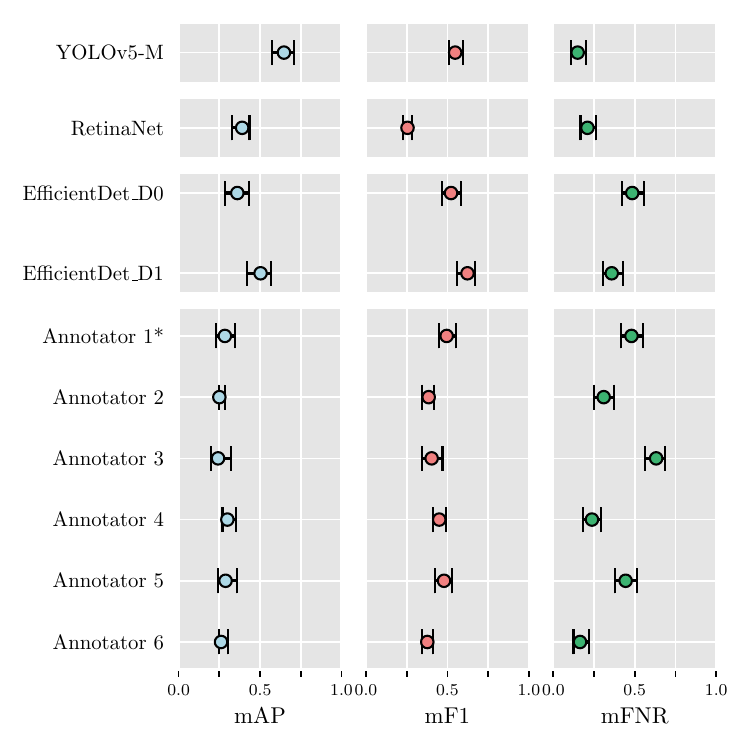}
    \caption{Bootstrap $95\%$ confidence intervals for the metrics mAP, mF1 and mFNR, for the models and the annotators}
    \label{fig:confidence_intervals}
    \end{minipage}
\end{figure}

The confidence intervals for the means of the distributions of the performance metrics, calculated for each model and each annotator, can be found in \Cref{tab:results_consensus_annotators_confint}. As described in \Cref{subsec:performance-evaluation}, these intervals were used to assess statistical significance between the different architectures, as well as between the models and the human expert rater performance. A graphical representation of these is show in \Cref{fig:confidence_intervals}, for ease of interpretation. 

Overall, the scores of all of the object detection models were similar to or better than that of the human expert annotators. In terms of AP, the YOLOv5 model achieved significantly higher scores than all of the annotators, as well as the RetinaNet and the EfficientDet D0. The RetinaNet and EfficientDet models also achieved mAP-scores that were similar to or significantly better than the annotators. Regarding F1, The YOLOv5 model achieved significantly higher scores than the RetinaNet model and 4 out of 6 annotators, but the difference with the EfficientDet models were not significant. The EfficientDet models achieved mF1-scores similar to or better than the annotators, whereas the mF1-score of the RetinaNet model was significantly lower than most of the annotators. In terms of FNR, the YOLOv5 model was significantly better (lower scores) than 4 annotators, and similarly, the mFNR of the RetinaNet was significantly better than 3 of the annotators. The EfficientDet models achieved mFNR scores that were similar to or significantly higher than the annotators, meaning that the performance was similar to worse than that of the annotators. \ReviewHighlight{The results per class can be found in Table 2 in Section 4 of the Additional Materials 1}.

    \begin{table}[ht]
    \centering
    \caption{Mean average precision (mAP), mean F1-score (mF1), and mean false negative rate (mFNR) evaluation of the models \ReviewHighlight{(trained on the fifth fold) }and individual annotators on the consensus test set, with an IoU-threshold of 0.3. All metrics are reported as score over the whole test set, and a $95\%$ confidence interval. \ReviewHighlight{The best results among the models and the annotators have been highlighted in bold letter.} * annotators who did not complete the individual annotation task (see \cref{subsec:validation-protocol}).} 
    \label{tab:results_consensus_annotators_confint}
    \begin{tabular}{r|cccc|cccc|cccc}
        \toprule
         Model / Annotator &   mAP &   mF1 &  mFNR\\
         \midrule
        YOLOv5, 5\textsuperscript{th} fold &  \textbf{0.647 [0.566, 0.707]} & \textbf{0.548 [0.506, 0.598]}& \textbf{0.149 [0.110, 0.203]}\\
         \midrule
         RetinaNet, 5\textsuperscript{th} fold & 0.407 [0.355, 0.458] & 0.177 [0.154, 0.202] & 0.210 [0.167, 0.262] \\
         \midrule
         EfficientDet D0, 5\textsuperscript{th} fold & 0.360 [0.290, 0.431] & 0.522 [0.461, 0.588]  & 0.484 [0.422, 0.552]\\ 
         EfficientDet D1, 5\textsuperscript{th} fold & 0.503 [0.421, 0.569] & 0.503 [0.421, 0.569] & 0.359 [0.306, 0.431]\\ 
         \midrule
         \midrule
        Annotator 1* &  0.284 [0.231, 0.347] &  \textbf{0.495 [0.447, 0.552]} & 0.480 [0.413, 0.552]\\
        Annotator 2 &  0.250 [0.247, 0.285]  &  0.385 [0.346, 0.420]  & 0.309 [0.251, 0.374]\\
         Annotator 3 & 0.242 [0.199, 0.320] & 0.403 [0.343, 0.470] & 0.631 [0.564, 0.686] \\
        Annotator 4 &  \textbf{0.299 [0.270, 0.353]} &  0.450 [0.411, 0.492] &  0.237 [0.180, 0.292]\\
        Annotator 5 &  0.288 [0.244, 0.356] &  0.479 [0.423, 0.528] & 0.444 [0.376, 0.515]\\
        Annotator 6 & 0.261 [0.248, 0.301]&  0.376 [0.346, 0.410]&  \textbf{0.164 [0.124, 0.217]}\\
        \bottomrule
    \end{tabular}
\end{table}
        
\section{Discussion}
\label{sec:discussion}
In the presented study, three different object detection DL architectures were trained and evaluated on the task of detection of proximal caries in BW X-ray images. The caries were annotated by dental clinicians and classified into three groups: enamel, dentine, and secondary lesions. The predictive performance of the models was assessed in terms of the object detection metrics AP, F1-score, and FNR, and compared against the performance of human expert annotators on a consensus test set. The main finding is that all model performances were on par with or better than the human annotators, with the best model achieving significantly higher scores than the human annotators for all metrics. 

The dataset presented in this study features $13,882$ BW images, with carious lesions annotated by six dental clinicians. To the best of our knowledge, this is the largest dataset presented so far for the task of training object detection models for caries detection, exceeding the size of the dataset described in \cite{Mayank2017} with $3,000$ images, and in \cite{Cantu2020} with $3,686$ BW images. A novel strategy for combining the annotations from multiple annotators on the same image was presented, creating robust ground truth annotations for training by combining the expert knowledge of all the annotators. In addition, a test set consisting of 197 images was jointly annotated by all the annotators by consensus agreement. The consensus test set was used to compare the model performances against the performance of the individual annotators, allowing for an assessment of the models usefulness by comparison against a baseline of human expert knowledge.

As detailed in \Cref{subsec:validation-protocol}, the performance of each of the architectures was assessed using five-fold cross validation. \ReviewHighlight{The folds were built through random sampling of the images without replacement. This approach could lead to data leakage as the split was performed at image level instead of patient level. Nonetheless, due to the size of the dataset, the augmentation during the pre-processing of the images, and the fact that those scans corresponding to a single patient show different regions of the denture, the risk of leakage is minimised. }In addition, all of the models were evaluated on the consensus test set, presented in \Cref{tab:results_AP_consensus_summary,tab:results_F1_consensus_summary,tab:results_FNR_consensus_summary}. The selected metrics, AP, F1-score, and FNR, were deemed appropriate for this experiment, as they summarise the goodness of the models to correctly identify the caries (AP), the trade off between precision and recall (F1-score), and the rate at which the object detectors disregard the caries which are in the BW images (FNR). By using the PASCAL VOC implementation of the metrics, the AP precision is regressed using a larger amount of points, compared to the 11-point interpolation of the AP curve, used in the COCO implementation of AP \cite{padilla2021}. This resulted in a better estimate of this metric, and was therefore considered adequate for this study. Lastly, to assess the statistical difference in performance of the models and the expert annotators, confidence intervals were estimated using the BCa algorithm \cite{boostrap1997}.

\begin{figure}[ht]
    \centering
    \includegraphics[width=0.9\textwidth]{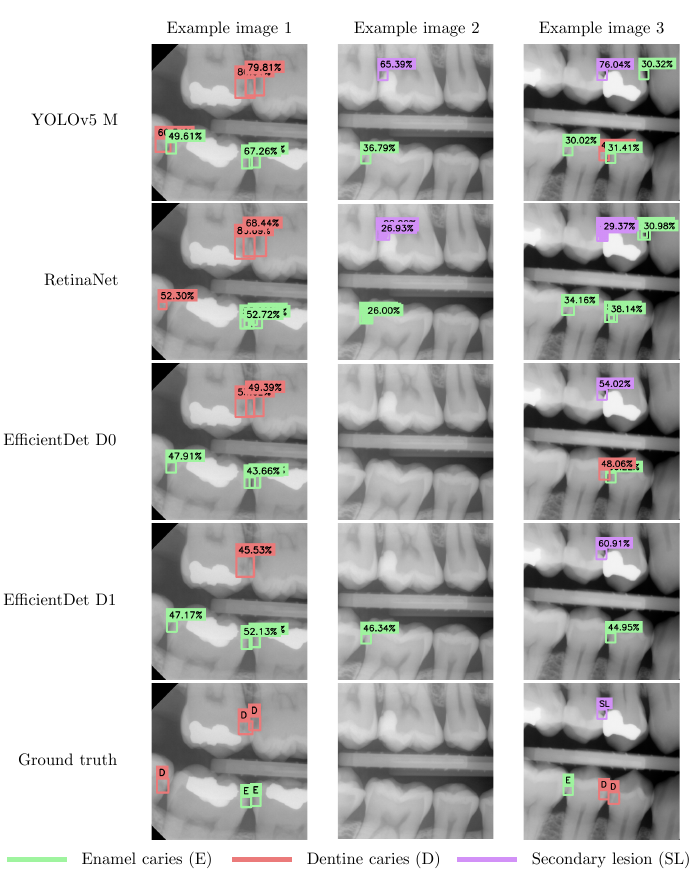}
    \caption{\ReviewHighlight{Detail of bitewing images from the consensus test set with predictions given by the trained models. The ground truth is shown in the bottom row.}}
    \label{fig:prediction_examples}
\end{figure}

The YOLOv5 model achieved the best performance in terms of the metrics used in the study. Both the EfficientDet D1 and YOLOv5 achieved significantly better performance than the RetinaNet in terms of mAP and mF1-score, even though the number of parameters for these models are lower than that of the of RetinaNet. Indeed, EfficientDet D1 is one fifth the size of RetinaNet, and yet it performed better in terms of mAP and F1. On the other hand, both the YOLOv5 and the RetinaNet achieved significantly lower FNR-scores than the EfficientDet models. In sum, all of the presented architectures exhibited different strengths and weaknesses, and an ensemble strategy of the models should be thus be considered, to improve the robustness of the predictions. \ReviewHighlight{\Cref{fig:prediction_examples} shows an example of the predictions given by each architecture on three different BW images, the ground truth is given for reference at the bottom row.}

Compared to equivalent previously published studies, comparable in scope with the presented work, the performances of the models are lower than the values reported in \cite{Devito2008, Berdouses2015, PradosPrivado2020, Schwendicke2019, Choi2018, Mayank2017, JaeHong2018, Shinae2021, Park2022}, although the values are not directly comparable as they are reported on different datasets. Unlike in these studies, the focus of this work was not to optimise and build a tailored object detection model, but to assess if the dataset was sufficient to obtain equivalent or better performance than dental clinicians, using state-of-the-art architectures. Indeed, as shown in \Cref{sec:results}, the trained models achieved significantly higher performances in sum on all of the metrics. A combination of the models strengths and weaknesses could thus be a solid foundation for an assistive tool for carious lesion detection in clinical practice.

As introduced in \Cref{sec:introduction}, the exclusive use of BW images to identify carious lesions is under-par, as it requires a follow-up direct inspection and probing of the infected area. However, the presented deep learning models have the potential to improve the efficiency of the analysis of the bitewing images and aid in the detection of these lesions, helping to speed up and improve the detection and diagnosis of caries.

The architectures included in this study were not modified nor tailored for the used dataset or applications, unlike previously published works \cite{Devito2008, Berdouses2015, Singh2017, Hwang2019, PradosPrivado2020, Schwendicke2019, Choi2018, Mayank2017, JaeHong2018, Shinae2021, Park2022}. Arranging the trained models in an ensemble fashion is expected to increase the overall performance, and the robustness of the predictions. Also, a patch-wise inference could further boost the performance by exposing the network to a closer view of the dental pieces, instead of working on the whole picture. Other augmentation techniques should be considered, such as gamma and brightness augmentations. \ReviewHighlight{Explainable AI techniques could be considered to better comprehend the decision process of the trained models, e.g., the features detected for each class}. Finally, future work should provide information regarding the inference runtime, so as to assess if it the detection models are suitable to be used in practice.

\section{Conclusions}
\label{sec:conclusion}
Detection and identification of caries on BW images entails several difficulties, including the monocular view of the dental structures, and hence, presence of artifacts due to the overlap of the dental pieces. Therefore, it is common practice to perform a visual inspection of the lesions found in the medical images. In this study, it has been shown how AI-powered object detectors can ease the task of finding these lesions in the images, with better performance than dental clinicians. To support this statement, three state-of-the-art object detection architectures were trained on the HUNT4 Oral Health Study BW image dataset, and evaluated against expert dental clinicians. Out of the three architectures, YOLOv5 (medium size) yielded the best results, achieving significantly higher scores than the expert annotators. A combination of the presented models can be used as an assistive tool in the clinic, to speed up and improve the detection rate of carious lesions. The usefulness of such a tool will be assessed in a future clinical validation study.


\section*{Abbreviations}
\label{sec:abbreviations}
\begin{tabular}{ll}
    \setlength{\tabcolsep}{12pt}
   \textbf{\textit{AI:}}  & Artificial intelligence \\
   \textbf{\textit{ML:}}  & Machine learning \\
   \textbf{\textit{DL:}}  & Deep learning\\
   \textbf{\textit{BW:}} & Bitewing image\\
   \textbf{\textit{OPG:}} & Panoramic X-ray image\\
   \textbf{\textit{IoU:}} & Intersection over union\\
   \textbf{\textit{NSM:}} &Non-maximum suppression algorithm\\
   \textbf{\textit{MDF:}} & Mixture density function\\
   \textbf{\textit{CLAHE:}}  & Contrast limited adaptive histogram equalization \\
   \textbf{\textit{AP:}} & Average precision\\
   \textbf{\textit{F1:}} & F1-score\\
   \textbf{\textit{FNR:}} & False negative rate\\
   \textbf{\textit{mAP:}} & Mean average precision across classes\\
   \textbf{\textit{mF1:}} & Mean F1-score across classes\\
   \textbf{\textit{mFNR:}} & Mean false negative rate across classes\\
\end{tabular}

\section*{Declarations}
\subsection*{Ethics approval and consent to participate}
Ethical approval has already been granted by the Regional Ethical Committee (REK) based in central Norway (project number 64645), and also had approval from Norsk Senter for Forskningsdata (reference number 718269). \ReviewHighlight{In the HUNT4 Oral Health Study, written and signed consent was acquired from the participants~\cite{HUNT4DataAnalysis}.}

\subsection*{Consent for publication}
Not applicable.

\subsection*{Availability of data and materials}
\ReviewHighlight{The HUNT Research Centre is authorised by the Norwegian Data Inspectorate to securely store and manage the data. De-identified data are shared with researchers upon approval of their research protocol by the Regional Ethical Committee and HUNT Research Centre. To safeguard participant privacy, HUNT Research Centre minimises data storage outside its data bank and refrains from depositing data in open repositories. Detailed records of all exported data for various projects are maintained in the HUNT data bank, and the centre can reproduce this information upon request. Data export is unrestricted, subject to approved applications submitted to HUNT Research Centre}. All the data in the this manuscript are available from TkMidt (contact: Abhijit Sen, \faEnvelope~abhijit.sen@ntnu.no) on reasonable request.

The code and trained models can be provided upon reasonable request to Boneprox~A.B. (contact: Shreya Desai, \faEnvelope~shreya.desai@boneprox.se).

\subsection*{Competing interests}
\ReviewHighlight{The authors declare the following financial interest/personal relationships that may be considered as
potential competing interests: SD is employee at Boneprox~A.B., and TR is CEO of Boneprox~A.S., and is co-founder of Boneprox~A.S.}


\subsection*{Funding}
The AI-Dentify project (project number 321408-IPNÆRINGSLIV20) is funded by the Research Council of Norway, under the scope of the Innovation Project for the Industrial Sector.

\subsection*{Authors' contributions}
\textbf{Conceptualization}: JPdF, RHH, SD, AS; \textbf{Methodology}: JPdF, RHH, SD, AS; \textbf{Data acquisition}: JPdF, LCN; \textbf{Formal analysis and investigation}: JPdF, RHH, SD, LCN; \textbf{Writing - original draft preparation}: JPdF, RHH, SD; \textbf{Writing - review and editing}: JPdF, RHH, SD, LCN, AS, TL; \textbf{Funding acquisition}: TL, TR, AS; \textbf{Resources}: TL, TR, AS; \textbf{Supervision}: TL, TR, AS; All authors read and approved the final manuscript.

\subsection*{Acknowledgements}
The authors would like to express their gratitude to the dental clinicians that helped with the annotations of the BW images: Trine Matheson Bye, Gunnar Lyngstad, Odd-Arne Opland, Harald Solem, and Mats Säll. Also to Theodor Remman, CEO of Boneprox~A.S., and project manager of the AI-Dentify project. Furthermore, we would like to thank Hedda Høvik, Astrid J. Feuerherm, and Patrik Cetrelli working at TkMidt for helping in data processing, logistics, and making resources available. 

\ReviewHighlight{The Trøndelag Health Study (HUNT) is a collaboration between HUNT Research Centre (Faculty of Medicine and Health Sciences, Norwegian University of Science and Technology NTNU), Trøndelag County Council, Central Norway Regional Health Authority, and the Norwegian Institute of Public Health.}


\bibliography{bibliography}

\end{document}